*Review Article*

# Clustering Approaches for Mixed-Type Data: A Comparative Study

**Badih Ghattas** [1] **and Alvaro Sanchez San-Benito** [2]

[1]*CNRS, AMSE, Aix-Marseille University, Marseille, France*
[2]*I2M, Airbus Helicopters, Aix-Marseille University, Marseille, France*

Correspondence should be addressed to Alvaro Sanchez San-Benito; alvaro.sanchez-san-benito@airbus.com

Clustering is widely used in unsupervised learning to find homogeneous groups of observations within a dataset. However, clustering mixed-type data remains a challenge, as few existing approaches are suited for this task. This study presents the state-of-the-art of these approaches and compares them using various simulation models. The compared methods include the distance-based approaches $k$-prototypes, PDQ, and convex $k$-means, and the probabilistic methods KAy-means for MIxed LArge data (KAMILA), the mixture of Bayesian networks (MBNs), and latent class model (LCM). The aim is to provide insights into the behavior of different methods across a wide range of scenarios by varying some experimental factors such as the number of clusters, cluster overlap, sample size, dimension, proportion of continuous variables in the dataset, and clusters' distribution. The degree of cluster overlap and the proportion of continuous variables in the dataset and the sample size have a significant impact on the observed performances. When strong interactions exist between variables alongside an explicit dependence on cluster membership, none of the evaluated methods demonstrated satisfactory performance. In our experiments KAMILA, LCM, and $k$-prototypes exhibited the best performance, with respect to the adjusted rand index (ARI). All the methods are available in *R*.

## 1. Introduction

Cluster analysis, or clustering, aims to classify data into groups (clusters) based on a measure of dissimilarity, so that data within clusters are as homogeneous as possible. As an unsupervised learning technique, clustering is commonly used in data analysis to extract insights from data and has application in various fields, including medicine, economics, or marketing. Real-world applications often lack a definitive approach to measure the quality of a clustering, leading to the development of various clustering techniques that depend on the specific problem and available data.

There are two primary categories of clustering techniques: hierarchical and nonhierarchical. The former groups similar data points together using an agglomerative or divisive process, which can be represented by a tree structure. In the contrary, the latter determines the specified number of clusters at once by iteratively maximizing (minimizing) a homogeneity criterion.

Most well-known clustering methods are based on dissimilarities or distance metrics, such as $k$-means [1]. These algorithms quantify the dissimilarity between data points using distance metrics such as Euclidean or Hamming distances. Another approach is to adopt model-based clustering [2], where data are assumed to be generated from a known mixture of distributions with unknown parameters. Indeed, data clustering can also be thought of as the process of inferring a probability distribution of a given dataset in the presence of hidden variables [3], in this case, cluster membership.

Currently, many real-world applications require the analysis of mixed-type data, which refer to datasets containing variables of different types (continuous, nominal, and/or ordinal). However, the research on clustering methods tailored to mixed-type data is still emerging (see the study by Ahmad and Khan [4] for a survey on the subject).

The simplest common approach consists of converting mixed-type data into a single type and using any clustering method for that type. However, this can result in information loss, especially when using distance-based clustering techniques, as discussed by Foss et al. [5]. Most existing direct approaches focus either on generalizing distance measures for mixed-type data or on using appropriate probabilistic models.

For the distance-based approaches' extension, the challenge lies in devising a dissimilarity measure that effectively accounts for each data type. Most methods use a convex combination of distances associated with each data type, such as, for instance, the Gower distance [6]. Such approaches differ in general in the choice of the data type weighting strategy; some algorithms such as $k$-prototypes [7] require a user-defined weight, while PDQ [8] considers the proportion of each variable type in the data. Convex $k$-means [9] automatically selects the optimal weights using a grid search that minimizes the ratio of within-cluster to between-cluster dispersion.

Regarding model-based methods, and despite their attractive statistical properties, they rely on strong parametric distributional assumptions. Among these approaches, KAymeans for MIxed LArge data (KAMILA) [10] presents a semiparametric model that combines $k$-means and mixture models without using a weighting strategy. Bayesian networks (BNs) appear to be a suitable approach for joint distribution estimation of mixed-type data and may be combined together with a mixture approach to achieve clustering.

There are very few benchmarking studies that compare clustering approaches for mixed-type data. Preud'homme et al. [11] compared some techniques from both distance-based and probabilistic approaches on simulated and real data. The experimental factors evaluated included sample size, the number of clusters, the ratio of continuous to categorical attributes in the data, the proportion of non-noisy variables, and the degree of relevance of the variables to the cluster structure. KAMILA, $k$-prototypes, and latent class model (LCM) [12] emerged as the top performing methods over partitioning around medoids (PAM), hierarchical clustering and LCA. Also, Jimeno et al. [13] conducted a simulation study that compares KAMILA, against $k$-prototypes and tandem analysis, based on two steps: a factor analysis for numerical encoding of the categorical variables, followed by a partitional clustering of the observations in the features space [14]. For the first step, multiple correspondence analysis (MCA) is used. For the clustering, the authors compared $k$-means, fuzzy $k$-means, probabilistic distance (PD) clustering, and Student-$t$ mixture models. Three experimental factors were examined: the number of clusters, the degree of clusters' overlap, and the ratio of continuous to categorical variables in the dataset. The results indicated that KAMILA and $k$-prototypes generally performed well for spherical clusters compared to the other techniques. The mixture of Student-$t$ distributions performed well for spherical and skewed clusters. More recently, a benchmarking study of distance-based methods by Costa et al. [15] compared KAMILA, $k$-prototypes, convex $k$-means, PAM (with Gower and Hening Liao weighting [16]), mixed $k$-means [17], and two tandem analysis techniques combining PCA for mixed data and $k$-means called FAMD/$k$-means[1] and mixed RKM[2] [18]. The authors used common experimental factors such as the number of clusters, the sample size, the proportion of categorical variables in the dataset, and the cluster overlap but also cluster distribution and sphericity. The results revealed KAMILA as the most effective method, followed by $k$-prototypes and sequential FAMD/$k$-means. In addition, the aforementioned studies used simulated data from multivariate Gaussian distributions.

Building on insights from previous works, we selected well-established methods known for their consistent performance across multiple studies (KAMILA, LCM, $k$-prototypes, and convex $k$-means). Furthermore, we incorporated more recent and promising approaches that have not yet been included in benchmarking studies (PDQ and BNs). These latter choices were motivated by their innovative clustering techniques or their ability to address large-scale clustering challenges. The study integrates methods from both distance-based and probabilistic model approaches, as outlined in Figure 1.

Based on suggestions from the previous works, we compare the aforementioned methods using datasets from four simulation models, including non-normal distributions and interactions between variables. In addition, we varied a wide range of experimental factors, including clusters' balance and size.

The remainder of the paper is structured as follows: Section 2 provides a short description of the relevant methods. Section 3 explains the simulation models and their configurations. Section 4 presents the results of the experiments. Section 5 presents the primary findings of the study and outlines open research questions related to the topic.

## 2. Clustering Methods for Mixed-Type Data

In this section, we presented a short overview of the state-of-the-art approaches which will be used later in our comparative study. All these approaches are nonhierarchical. They may be classified in two subgroups, those optimizing distance-based criteria ($k$-prototypes, PDQ, and convex $k$-means) and those using probability-based criteria (KAMILA, LCM, and MBN).

We introduced some notations before presenting the methods. We denoted random variables with capital letters, realizations of these variables with small letters, and vectors in bold font. Let $\mathbf{V}$ denote a random $R$-dimensional continuous vector with coordinates $V_r \in \{1, \ldots, r, \ldots, R\}$, and $\mathbf{W}$ denotes a random $S$-dimensional discrete vector with coordinates $W_s \in \{1, \ldots, s, \ldots, S\}$. When necessary, $\mathbf{W}$ can be split into $\mathbf{W^N}$ and $\mathbf{W^O}$ (nominal and ordinal components), with dimensions $S_N$ and $S_O$, respectively. We have a sample of $n$ random independent variables identically distributed as $\mathbf{X} = (\mathbf{V^T}, \mathbf{W^T})$ in dimension $(R + S) \times 1$, denoted $\mathbf{X}_i$ with $i \in \{1, \ldots, n\}$. Finally, $X_{ij}$ is the $i^{\text{th}}$ observation of the $j^{\text{th}}$ component of **X**. A similar notation with small letters is used to denote the realizations of these variables: $x$, $x_i$, and $x_{ij}$.

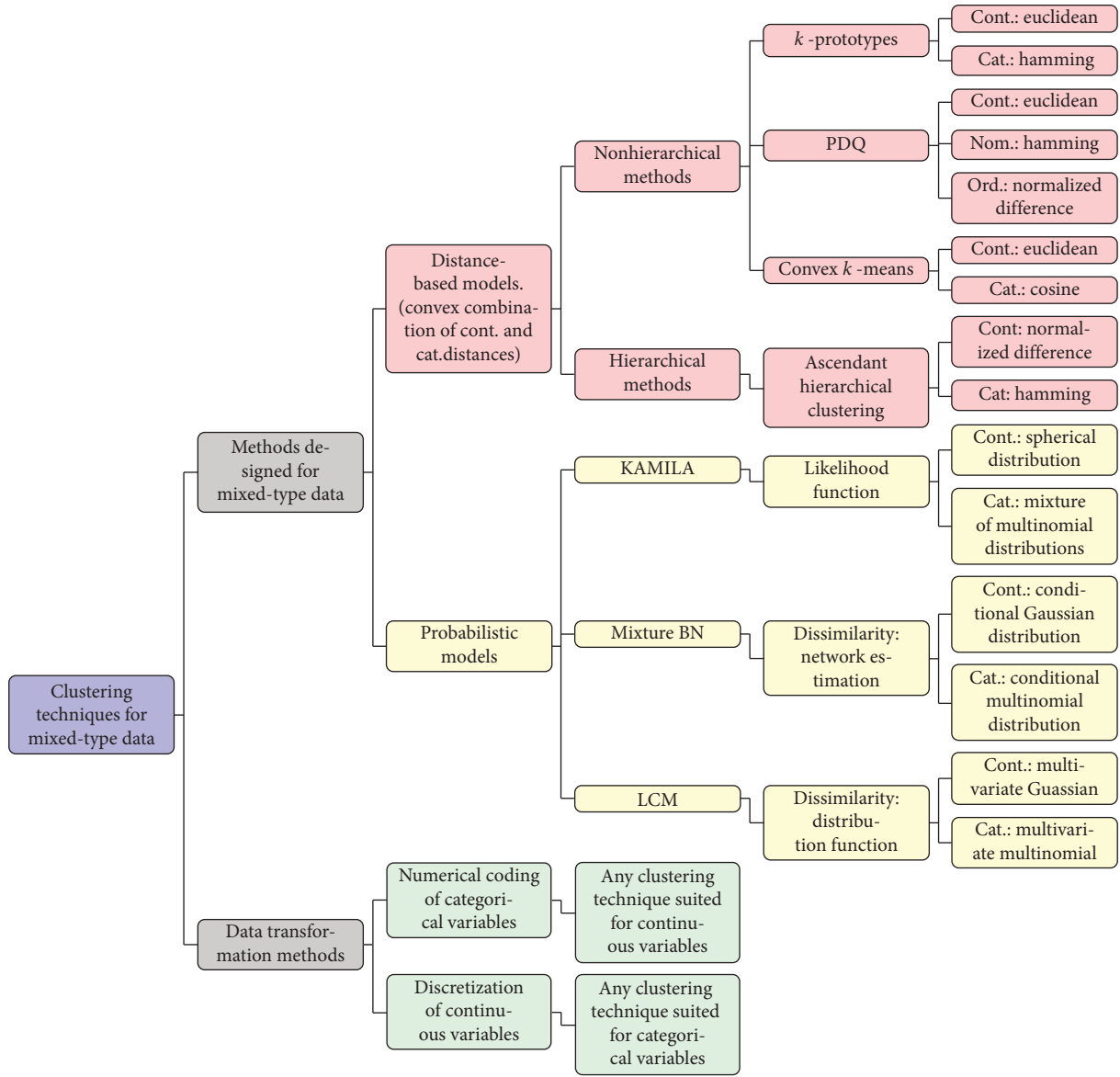

Figure 1: Clustering mixed-type data approaches.

### 2.1. Distance-Based Approaches.

All approaches described in this section share the same principle used by *k*-means, initializing centers and alternating a partitioning (assignment) step and an update step of the centers. They differ mainly in the design of the distance used to assign an observation to a cluster.

*K-prototypes* [7] is an iterative approach that combines *k*-means and *k*-modes. It uses an hybrid distance function involving the squared Euclidean distance for continuous attributes and the Hamming distance for categorical attributes. The latter's contribution to the distance function is balanced through a single weight $\gamma$ that does not depend neither on the categorical variables nor on their number. The clustering process follows a similar procedure as that of *k*-means taking $K$ as input, but it also considers the hyperparameter $\gamma$, defined by the user. Several weighting strategies are proposed in the literature ([19, 20]). *k*-prototypes minimizes the following objective function:

$$P(\boldsymbol{\mu}) = \sum_{k=1}^{K} \sum_{i=1}^{n} \mathbf{1}_{ik} \left( \sum_{j=1}^{R} (x_{ij} - \mu_{kj})^2 + \gamma \sum_{j=R+1}^{R+S} \mathbf{1}_{(x_{ij} \neq \mu_{kj})} \right), \quad (1)$$

where $\boldsymbol{\mu}_k = \{\mu_{k,1}, \ldots, \mu_{k,R}, \mu_{k,R+1}, \ldots, \mu_{k,R+S}\}$ is the center of Cluster $k$, and $\mathbf{1}_{ik} = 1$ if $i$ is in Cluster $k$ and $\mathbf{1}_{ik} = 0$ otherwise. The $R$ first components of $\mu_k$ correspond to the sample mean for the continuous attributes and the $S$ last components correspond to the sample mode for the categorical attributes within Cluster $k$ and $\boldsymbol{\mu} = \{\boldsymbol{\mu}_k, k = 1, \ldots, K\}$.

*K-prototypes* extends *k*-means approach to mixed-type data by unifying the criteria used for continuous and categorical variables. Nonetheless, the challenge lies in selecting the optimal value for the parameter $\gamma$.

*PD Clustering (PDC)* [21] is a two-step method that takes as input the number of cluster $K$ and iteratively updates the centers and the membership of clusters. It minimizes a classifiability criterion called the joint distance function (JDF), which depends on the distances $d_{ik}$ of each data point to the centers, and on the probability $p_{ik}$ of each point belonging to a cluster.

$$\text{JDF} = \sum_{i=1}^{n} \sum_{k=1}^{K} p_{ik} d_{ik}. \tag{2}$$

Tortora and Palumbo [8] proposed an extension of the PDC adjusted for cluster size (PDQ) to mixed-type data. Furthermore, it takes into account the quantity of nominal variables, denoted as $S_N$, and ordinal variables, $S_O$, along with the number of continuous variables, $R$, in the dataset. To determine the dissimilarity between data points and their respective centers, Euclidean distance is utilized for continuous variables, normalized difference for ordinal variables, and Hamming distance for nominal variables. Gower's dissimilarity is used to incorporate the three distinct dissimilarity measures into a single distance. Thus, the JDF becomes

$$JDF = \sum_{i=1}^{n} \sum_{k=1}^{K} \frac{p_{ik}^2}{s_k} \left( \alpha_C d_{ik}^C + \alpha_O d_{ik}^O + \alpha_N d_{ik}^N \right), \tag{3}$$

where $\alpha_C = R/p$, $\alpha_O = S_O/p$, $\alpha_N = S_N/p$ and $p = R + So + Sn$, $d_{ik}^C = \sqrt{\sum_{j=1}^{R}((x_{ij} - \mu_{kj})/x_j^*)^2}$, and $x_j^* = 1$ if $-0.1 < \overline{x}_j < 0.1$, $x_j^* = \overline{x}_j$, otherwise, where $\overline{x}_j$ is the empirical mean of the variable $j$, $d_{j=1}^{N} = \sum_{j=R+S_O+1}^{p} \mathbf{1}_{(x_{ij} \neq \mu_{kj})}$ and $d_{ik}^O = \sum_{j=R+1}^{R+S_O} |x_{ij} - \mu_{kj}|/R_j$ where $R_j$ stands for the range of the $j^{\text{th}}$ variable, and $s_k$ stands for the size of Cluster $k$. The centers of each cluster are specific for each data type and computed as $\partial JDF/\partial \mu_{kj} = 0$ for $j = 1, \ldots, p$. At each iteration, PDQ computes the dissimilarity and probability matrices and updates the cluster centers accordingly.

PDQ distinguishes itself by separately considering nominal and ordinal variables and offering a probabilistic membership rather than a crisp membership (where a point belongs or does not belong to a given cluster).

*Convex k-means* is an extension of *k*-means to cluster mixed-type data within a similar framework to *k*-prototypes, that is, estimating a weight $\alpha$ to determine the relative contribution of each data type. The authors present a general framework for multitype data, with a primary focus on mixed data that include both continuous and dummy coded categorical variables. They first defined a weighted dispersion measure for mixed data as

$$D^{\alpha}(\mathbf{x}_1, \mathbf{x}_2) = \alpha_{\text{cont}} D_{\text{con}}(\mathbf{v}_1, \mathbf{v}_2) + \alpha_{\text{cat}} D_{\text{cat}}(\mathbf{w}_1, \mathbf{w}_2), \tag{4}$$

where $D_{\text{con}}$ is the squared-Euclidean distance, $D_{\text{cat}}$ is the cosine distance, and $\sum_{l=1}^{2} \alpha_l = 1$ for $\alpha = (\alpha_{\text{cont}}, \alpha_{\text{cat}})$. For a fixed weighting $\alpha$ and $K$, convex *k*-means partitions the data as follows:

$$\{\mathscr{P}_k^*\}_{k=1}^{K} = \underset{\{\mathscr{P}_k\}_{k=1}^{K}}{\text{argmin}} \left( \sum_{k=1}^{K} \sum_{x \in \mathscr{P}_k} D^{\alpha}(x, \mu_k) \right), \tag{5}$$

where $\mu_k = \{\mu_{k,\text{cont}}, \mu_{k,\text{cat}}\}$ denotes the center of Cluster $k$. To determine the optimal partition $\{\mathscr{P}_k^*\}_{k=1}^{K}$, convex *k*-means follows the same approach as *k*-means.

In order to identify the optimal weighting $\alpha^*$, the authors define the average within-cluster dispersion for each data type as

$$\begin{aligned} \mathscr{W}_{con}(\alpha) &= \sum_{k=1}^{K} \sum_{\mathbf{v} \in \mathscr{P}_k^*(\alpha)} D_{\text{con}}(\mathbf{v}, \mu_{k,\text{con}}^*(\alpha)), \\ \mathscr{W}_{cat}(\alpha) &= \sum_{k=1}^{K} \sum_{\mathbf{w} \in \mathscr{P}_k^*(\alpha)} D_{\text{cat}}(\mathbf{w}, \mu_{k,\text{cat}}^*(\alpha)), \end{aligned} \tag{6}$$

and the average between-cluster dispersion as

$$\begin{aligned} \mathscr{B}_{con}(\alpha) &= \sum_{i=1}^{n} D_{\text{con}}(\mathbf{v}_i, \overline{\mu_{\text{con}}}) - \mathscr{W}_{con}(\alpha), \\ \mathscr{B}_{cat}(\alpha) &= \sum_{i=1}^{n} D_{\text{cat}}(\mathbf{w}_i, \overline{\mu_{\text{cat}}}) - \mathscr{W}_{cat}(\alpha), \end{aligned} \tag{7}$$

where $\overline{\mu_{\text{cont}}}$ and $\overline{\mu_{\text{cat}}}$ are the centers computed across all continuous and categorical variables.

Finally, to select the optimal weighting $\alpha^*$, convex *k*-means conducts repeated clustering across a grid of values and identifies the $\alpha$ that minimizes $Q$, such that

$$\mathscr{Q}(\alpha) = \frac{\mathscr{W}_{con}(\alpha)}{\mathscr{B}_{con}(\alpha)} \times \frac{\mathscr{W}_{cat}(\alpha)}{\mathscr{B}_{cat}(\alpha)}. \tag{8}$$

*Convex k-means* [9] originally proposes the objective function as a generalized Fisher's ratio. However, it presents a significant drawback: When the number of unique combinations of categorical levels (e.g., two binary variables with $2 \times 2$ level combinations) matches the number of cluster $k$, each distinct level combination is assigned to its individual cluster, resulting in an average within-cluster dispersion value of zero for the categorical variables. Consequently, the clustering process entirely disregard the continuous variables.

### 2.2. Probabilistic Approaches.
The methods outlined in this section are based on distributional assumptions for the data.

*KAMILA* [10] is an iterative semiparametric approach that combines *k*-means and mixture models. Continuous variables follow a mixture of spherical distributions of density $h$ such that $\mathbf{V} \sim f_\mathbf{V}(\mathbf{V}) = \sum_{k=1}^{K} \pi_k h(\mathbf{v}; \mu_k)$ where $K$ is the number of clusters, $\pi_k$ is the mixture coefficient, and $\mu_k$ are the centers of the $k$-th cluster. For spherical symmetric distributions, this multidimensional density $f_\mathbf{V}(\mathbf{V})$ is shown to be proportional to a univariate density $f_V(r)$ over the distance $r = \sqrt{v^T v}$ to the center and is estimated using classical kernel estimator denoted as $\hat{f}_V$. Categorical variables follow a mixture of multinomial distributions such that $W \sim f_W(W) = \sum_{k=1}^{k} \pi_k \prod_{s=1}^{s} m(w_s; \theta_{ks})$ where $m(.;.)$ stands for the multinomial probability mass function and $\theta_{ks}$ for the parameters of the $s$-th variable drawn from the $k$-th cluster. It is assumed that categorical variables are independent within each Cluster $k$.

The number of Cluster $K$ is fixed by the user. First, the centers of the clusters and the parameters of the multinomial distribution are initialized randomly using appropriate distributions and supports. Each iteration in KAMILA comprises of a partitioning step and an estimation step. Each observation $i$ is assigned to one of the $k$ clusters by maximizing the following objective function

$$H_i(k) = \log[\hat{f}_V(d_{ik})] + \log P[W_{ik} = w_{ik}], \quad (9)$$

where $W_{ik}$ is the restriction of the discrete variables to cluster $k$, $d_{ik}$ stands for the Euclidean distance between the observation $i$ and the cluster center $\mu_k$, and $P$ is the multinomial probability mass function. The estimation step updates the new centers and estimates the parameters of the multinomial probability function as well as the density for the continuous variables.

The novelty of KAMILA's approach lies in the use of likelihood as an objective function and in avoiding the computation of a multivariate density estimate using a univariate density equivalent for spherical distribution. Also, KAMILA does not use a weighting to balance the contribution of continuous and categorical variables.

*Mixture of BN (MBN)*. Pham and Ruz [22] suggested a framework for training BN classifiers in an unsupervised manner for clustering purposes. Data are assumed to be generated by a mixture of $K$ BN,

$$P(X) = \sum_{k=1}^{K} \alpha_k f_k(X), \quad (10)$$

where $\alpha_k$ is the mixing coefficient satisfying $\sum_{k=1}^{k} \alpha_k = 1$ and $f_k$ is the mixing component distribution. The parameters of this model are denoted as $\theta$, and they include $\alpha_k$ and the parameters of $f_k$. Clustering the data points is done through the estimation of the unknown parameters $\theta$. To do that, the classification maximum likelihood (CML) criterion [23] is maximized as follows:

$$\text{CML}(\theta|X) = \sum_{k=1}^{K} \sum_{x \in \pi_k} \log f_k(x) + \sum_{k=1}^{K} n_k \log \alpha_k, \quad (11)$$

where $\{\mathcal{P}_k\}_{k=1}^{K}$ is a partition of the $n$ data points and $n_k$ is the number of individuals in $\mathcal{P}_k$. The CML may be maximized using the classification expectation-maximization (CEM) [24] algorithm, which is an adaptation of the EM algorithm for clustering purposes. From an initial random partition $\{\mathcal{P}_k^0\}_{k=1}^{K}$, and a fixed value of $K$, each iteration $t$, for $t > 0$, consists of the following:

1. The $E$-step computes the probabilities that observation $i$ belongs to Cluster $k$ for $i = 1, \ldots, n$ and $k = 1, \ldots, K$, using the actual estimate of $f_k$
2. The $C$-step updates and computes a new partition by assigning each observation $i$ to the cluster that provides the highest probability, setting a new partition $\{\mathcal{P}_k^t\}_{k=1}^{K}$
3. The $M$-step computes the maximum-likelihood estimates of $\theta_k^t$ given $\mathcal{P}_k^t$ for $k = 1, \ldots, K$

Pham and Ruz [22] applied the above procedure using three different BN classifiers for $f_k$; Chow and Liu multinet classifiers, tree-augmented Naive Bayes, and the simple BN classifiers. BN classifiers use BN to model the probabilistic relationships between features and class labels, enabling predictions based on these relationships. These supervised BN classifiers are only available for discrete variables. To use the CEM approach for mixed-type data, we replace the BN classifiers by a classical unsupervised BN algorithm. The advantage of a mixture of BN is that it accounts for dependences between variables for the clustering.

*LCM* [12] is a mixture model that can also identify important variables for the clustering as well as the optimal number of clusters. Mixture models assume that continuous variables follow a multivariate normal distribution, whereas categorical variables follow a multivariate multinomial distribution. In addition, it is assumed that variables are independent within clusters. A variable $j$ is considered irrelevant for clustering if its distribution is similar across all clusters. Let $\Omega$ be the set of relevant variables. For a mixture model with $K$ components, the probability density function for an observation $x_i$ is defined as follows:

$$f(x_i|\theta) = \prod_{j \in \Omega^c} h_j(x_{ij}|\theta_{1j}) \sum_{k=1}^{K} \alpha_k \prod_{j \in \Omega} h_j(x_{ij}|\theta_{kj}), \quad (12)$$

where $h_j(x_{ij}|\theta_{1j})$ is the distribution function for an irrelevant variable and $h_j(x_{ij}|\theta_{kj})$ is the distribution function for Cluster $k$ with parameters $\theta_{kj}$. Also, $\alpha_k \in [0, 1]$ are the mixing coefficients, satisfying $\sum_{k=1}^{K} \alpha_k = 1$. The set of parameters of the model is $\theta = \{\alpha_1, \ldots, \alpha_k, \theta_{11}, \ldots, \theta_{kj}\}$.

Similar to classical mixture models, the optimal partition is obtained via the EM algorithm. This iterative algorithm alternates between two steps: computing $\theta$ to maximize the log-likelihood and updating the partition based on the conditional expectation of the complete-data log-likelihood. If variable selection is requested, a penalization on the BIC or MICL criterion is applied during the maximization step. In the experiments conducted in Section 3, the number of Cluster $K$ is provided and variable selection is not enabled, as it is beyond the scope of this work. However, these functionalities may be of interest to users and are therefore worth mentioning.

All the approaches presented in this section are accessible for use in *R*.

## 3. Experiments

In this section, we described the simulation models, preprocessing strategies, and software utilized to evaluate and compare the performance of the clustering methods described in the previous section.

### 3.1. Simulation Models.
We generated synthetic mixed datasets from four main models, varying the following experimental factors: number of clusters ($K$), cluster distribution ($\pi^*$), cluster overlap, sample size ($N$), ratio of continuous to categorical variables, and total number of variables ($p$). Among these, the number of clusters, cluster overlap, sample size, and the ratio of continuous to categorical variables are commonly examined in simulation studies and have been identified as critical factors influencing clustering performance. In contrast, cluster distribution is less frequently considered in the literature but has been highlighted as a valuable aspect for the analysis by Preud'homme et al. [11]. We believe that the experimental factors included in this study are representative of real-world datasets and can provide guidance to practitioners in selecting the most suitable clustering approach according to their case.

#### 3.1.1. M1: Multivariate Gaussian Model.
Following the methodology of McParland and Gormley [25] and Tortora and Palumbo [8], the continuous and the categorical variables were simulated from a mixture of multivariate Gaussian densities, with dimensions corresponding to the number of variables and covariance matrix as the identity matrix. The degree of overlap was controlled by manipulating the means of individual variables within each cluster. Specifically, for continuous variables, the means for Cluster 1 were obtained using the sequence from 0 to 10 with a step of $10/(R-1)$ where $R$ is the desired number of continuous variables. The means of the observations belonging to the other clusters were computed as $\mu_{k+1} = \mu_k + 5 - (\text{overlap} * 5)$ for each continuous variable and $k = 1 \ldots, K$. To simulate two continuous variables for three clusters with 20% overlap, the means would be (0, 10), (4, 14), and (8, 18). For categorical variables, an underlying continuous vector with dimension $m-1$ is generated from a multivariate Gaussian distribution, where $m$ stands for the possible number of categories of a discrete variable. The observed categorical response is determined by comparing elements of the continuous vector to a threshold. If the maximum element is less than 0, the response is 1. Otherwise, it is determined by the index of the maximum element greater than 0. From this model, we generated 36 different datasets (unique combinations of experimental factors) in Dimension 12, varying in number of clusters (2, 5, and 10), cluster overlap (30% and 60%), cluster size (700 and 1400), and proportion of continuous variables in the dataset (33%, 50%, and 66%).

#### 3.1.2. M2: Exponential-Discrete Model.
This simulation model is inspired by a real-world dataset from the aerospace industry. Datasets generated from this model follow an exponential-discrete distribution. All simulated clusters follow the same distributions but with varying parameter combinations.

Continuous variables are distributed according to a mixture of exponential distributions with the following pdf:

$$g(x; \lambda^{(k)}) = \begin{cases} \lambda_1^{(k)} e^{-\lambda_2^{(k)} x} - \lambda_3^{(k)} e^{-\lambda_4^{(k)} x}, & \text{if } x \geq 0, \\ 0, & \text{if } x < 0, \end{cases} \quad (13)$$

where $\lambda^{(k)} = \{\lambda_1^{(k)}, \ldots, \lambda_4^{(k)}\}$ and $k = 1, \ldots, K$. Nominal variables follow a discrete distribution with probabilities $\mathbf{p}_k \in [0,1]^m$ where $m$ is the number of modalities of the nominal variables, fixed in our simulations to 14. Finally, binary variables follow a Bernoulli distribution $\text{Bern}(p_k)$.

From this model, we generated datasets from 24 experimental scenarios for each value of $K$. Table 1 gives a description of all the scenarios and the parameters used for the simulations. Figure 2 shows the scatter plot of a synthetic dataset generated from M2.

#### 3.1.3. M3: BN Classifier Model.
The number of clusters is fixed to $K$. The datasets are simulated using the SBN classifier with the network of Figure 3(a). This simulation model introduces strong interactions between the variables with explicit dependence with the cluster membership. The model consists of one multinomial variable ($C$), three conditional multinomial variables ($X_1, X_2$, and $X_5$) and three conditional Gaussian variables ($X_3, X_4$, and $X_6$). The variable $C$ denotes cluster membership and follows a uniform distribution over the $K$ levels, thus $P(C = k) = 1/K$ for $k = 1, \ldots, K$. The discrete variables $X_1$ and $X_2$ have three levels, and $X_5$ is binary. The conditional probability values of the discrete variables are fixed by a permutation of the set {0.64, 0.33, 0.04} for variables $X_1$ and $X_2$, and of the set {0.1, 0.9} for variable $X_5$. The $3 * K$ conditional means of the variable $X_3$ are fixed using the sequence from 0.5 to $1.5 * K$ with a step of 0.5, and for variable $X_4$, using the sequence from $2 * K$ to $5 * K - 1$ with a step of 1. Finally, the $K$ conditional means of the variable $X_6$ are fixed using the sequence from $5 * K - 1$ to $6 * K - 2$ with a step of 1. To fix the standard deviation of each conditional Gaussian variable, we randomly sampled a value from the set $\{0.5, 0.6, \ldots, 1.5\}$. Figure 2 shows the scatter plot of the simulated dataset for 500 instances. For $K$, we tested the values 2, 4, and 6 and we varied the sample size $N$ to 300 or 1200.

#### 3.1.4. M4: Mixture of BNs.
For a fixed number of Cluster $K$, we generated samples coming from a mixture of $K$ BNs, having all the same structure, as shown in Figure 3(b), but with different conditional probability tables. The clusters have the same size, thus $\alpha_k = 1/K$ for $k = 1, \ldots, K$. In this model, we have also interactions between the variables, with implicit dependence to the cluster membership. The variables $X_1, \ldots, X_6$ are of the same type as in the simulation model M3. The means of the conditional Gaussian variables $X_3$ and $X_4$ are fixed using the sequence from $(c + 2 * k)$ to $c + 2 * k + 2$ with a step of 1, with $c = 0$ for $X_3$ and $c = 5$ for

TABLE 1: Description of the simulation study of the exponential-discrete model (M2).

$N = 300, 1200$
% Cont. variables = 33%, 50%, and 66%
Nb. binary var. = 50% of cat. variables
Dimension = 6, 12
$m = 14$

| Clusters' distribution | $K = 2$<br>$(1/2 − 1/2), (1/5 − 4/5)$ | $K = 3$<br>$(1/3 − 1/3 − 1/3), (1/2 − 1/4 − 1/4)$ |
|---|---|---|
| $k = 1$ | | |
| $\lambda$ | (1.3, 0.05, 20, and 1) | (3.5, 0.05, 20, and 1) |
| $\mathbf{p}_1$ | (0.5, 0.02, 0.013, 0.03, 0.02, 0.02, 0.017, 0.01, 0.01, 0.06, 0.1, 0.08, 0.05, and 0.07) | (0.25, 0.2, 0.15, 0.1, 0.1, 0.05, 0.03, 0.02, 0.02, 0.02, 0.02, 0.02, 0.01, and 0.01) |
| $p_1$ | 0.64 | 0.8 |
| $k = 2$ | | |
| $\lambda$ | (1.1, 0.05, 20, and 1) | (1.3, 0.05, 20, and 1) |
| $\mathbf{p}_2$ | (0.08, 0.02, 0.08, 0.13, 0.05, 0.03, 0.12, 0.05, 0.01, 0.15, 0.01, 0.2, 0.03, and 0.04) | (0.01, 0.01, 0.02, 0.02, 0.02, 0.02, 0.02, 0.03, 0.05, 0.1, 0.1, 0.15, 0.2, and 0.25) |
| $p_2$ | 0.3 | 0.5 |
| $k = 3$ | | |
| $\lambda$ | | (1.1, 0.05, 20, and 1) |
| $\mathbf{p}_3$ | | (0.01, 0.02, 0.02, 0.05, 0.1, 0.15, 0.25, 0.15, 0.1, 0.1, 0.02, 0.01, 0.01, and 0.01) |
| $p_3$ | | 0.2 |

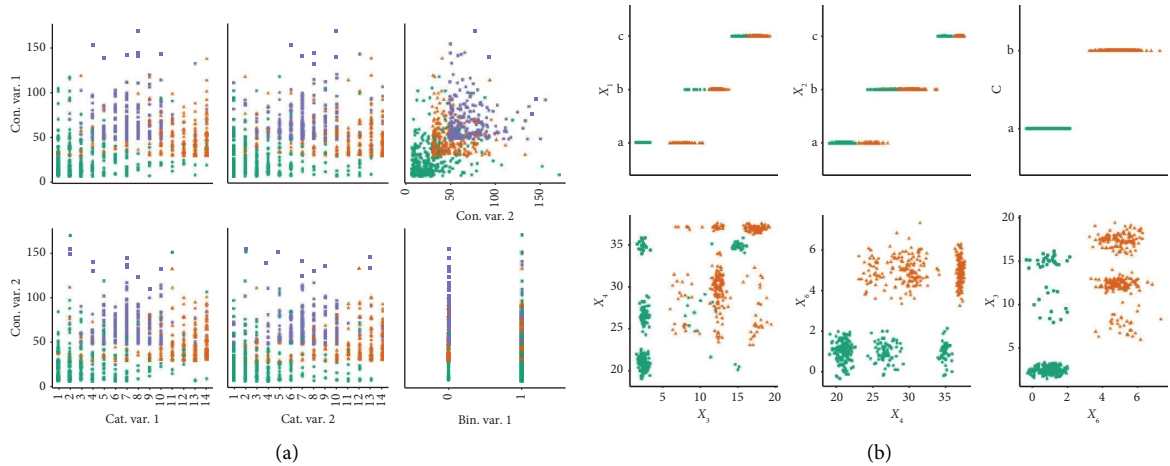

Figure 2: Scatter plot of the simulation models M2 (exponential-discrete) and M3 (BN classifier). Color and dot shape represent cluster membership. (a) M2: two continuous and three categorical variables with $K = 3$ and $N = 700$; (b) M2: three continuous and three categorical variables with $K = 2$ and $N = 500$.

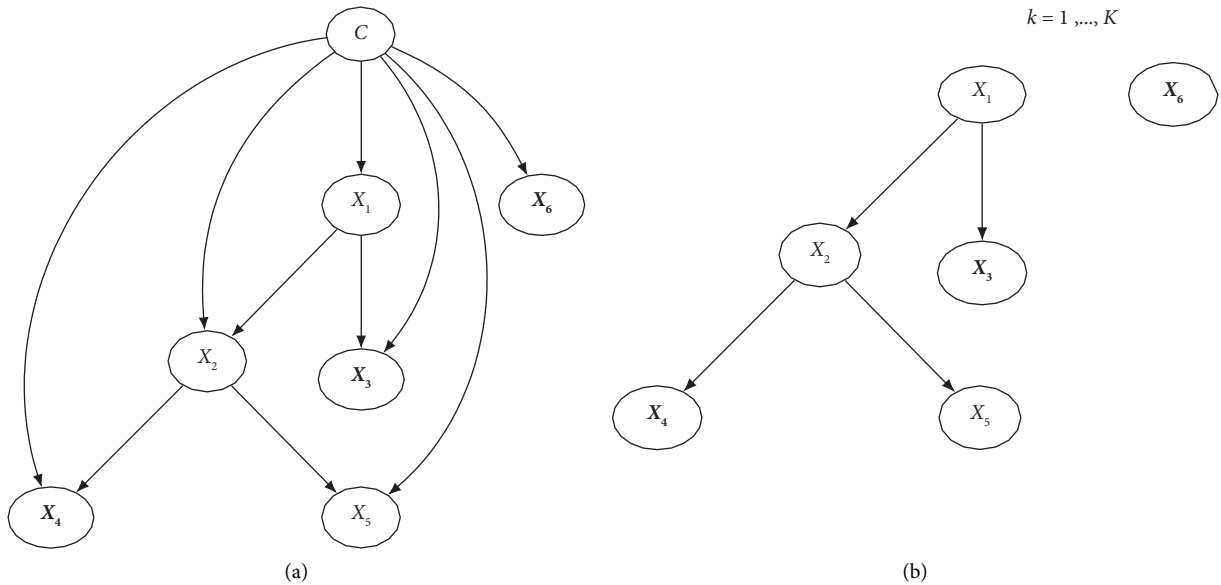

Figure 3: Network representations of simulation models M3 (BN classifier) and M4 (mixture of BN). Bold font denotes continuous variables. (a) M3 model representation. (b) M4 model representation.

$X_4$. The conditional mean of variable $X_6$ is fixed to $2 * k + 9$. The standard deviations for the conditional Gaussian variables and the conditional probability values of the discrete variables are fixed according to the M3 strategy. For $K$, we tested the values 2, 4, and 6 and we varied the simple size $N$ to 300 or 1200.

3.2. Data Preprocessing and Evaluation Criteria. Out of all the chosen methods, only convex $k$-means requires binary transformation of the categorical variables. To assess the clustering results across methods that optimize different criteria and facilitate comparison with other studies, the Adjusted Rand Index (ARI) was used. The ARI [14] is a measure based on pairwise agreements that corrects the Rand Index for chance. A value of one indicates perfect class agreement, while a value of 0 denotes random classification. In addition, adjusted mutual information (AMI) [26] was also calculated to evaluate clustering performance for the M2 simulation model, as it is expected to be less sensitive to cluster size imbalance compared to ARI [27]. To achieve stable results, we analyzed each combination of experimental factors over 10 replicates. We reported the mean ARI scores for each simulation model across replicates.

3.3. Software and Tuning. All the analyses in this paper were performed using *R* statistical software (version 4.0.1; *R* Core Team) [28]. For the $k$-prototypes method, we used the kproto function from the package clustMixType [29]. To ensure consistent results, we set the number of random initialization to 10. The value of the distance parameter $\gamma$ was automatically determined by the function lambdaest (in the same package) as the ratio of averages over all continuous/categorical variables. Convex $k$-means method was performed using the function gmsClust from the package KAMILA [30]. The optimal weighting value for the continuous variables is selected from a grid of 20 values equally spaced in the interval (0, 1). The next method we considered is KAMILA, using the KAMILA function (from the package of the same name). To guarantee stable results, we fixed the number of random initializations to 20. For the PD clustering adjusted for cluster size, PDQ, we used the function PDQ from the package FPDclustering [31]. We kept the default method $k$ medoid to initialize cluster centers, and the distance measure used was Gower. For LCM, we used the function VarSelCluster from the package VarSelLCM [32] developed by the authors. The last method we considered is the mixture of BN. To learn the structure from the data and compute the posterior probabilities within the CEM algorithm, the function tabu from the package bnlearn [33] was used with the score BIC to optimize. To reduce the chances of converging to a local optimum, 10 random initializations were chosen along with an S-step of 200 iterations. Data from Model M1 were generated using the function *mix_data* from Roy et al. [34]. The code used for the analysis and some results omitted in the paper are publicly available at https://github.com/Alsanchez13/Clustering-approaches-for-mixed-type-data-A-comparative-study.

## 4. Results and Discussion

Results from the multivariate Gaussian simulation model (M1) are depicted in Figure 4. The performance of the methods under consideration is mainly determined by the number of clusters and the degree of cluster overlap, with the exception of the MBN. In all scenarios, LCM, KAMILA, and convex $k$-means exhibit similar behavior, with ARI values of one or close. The behavior of $k$-prototypes is similar to that of the previously mentioned methods. However, it exhibits a greater decrease in clustering accuracy as the number of clusters and the degree of cluster overlap increase. The performance of PDQ is less affected by an increase in the number of clusters compared to other methods, showing an almost perfect clustering when $K$ is equal to 10 and with a 30% overlap. However, it is more vulnerable to a higher degree of cluster overlap and a higher proportion of categorical variables in the dataset. Surprisingly, it shows the worst ARI values when $K$ is equal to two and a 60% overlap. Out of all the methods studied, MBN shows the poorest performance. It is particularly vulnerable to small sample sizes and tends to work better when there are fewer continuous variables in the dataset. Findings regarding the influence of the degree of overlap and the proportion of continuous variables in the dataset are inconclusive for this method.

Model M1 is well-suited for LCM as well as for $k$-means-based methods, as the clusters are spherical and moderately well separated, and the variables are independent. Therefore, the performance of LCM, KAMILA, $k$-prototypes, and convex $k$-means is not surprising. The MBN only demonstrates good performance when correctly identifies the true underlying data structure, that is, for $K = 5$ and a 60% overlap.

The performance of the clustering methods on the exponential-discrete model (M2) depends on various factors. Unlike the M1 model, the shape of the clusters in M2 are not spherical, as depicted in Figure 2.

For $K = 2$, Table 1 shows the impact of unbalanced cluster distribution on the clustering performance. As expected, the clustering results generally improve as $N$ and the dimension increase. When cluster distribution is balanced, KAMILA emerges as the top-performing method (Table 2). Specifically, in this setup, Cluster 1 is characterized by Level 1 with a probability of 50%. $k$-prototypes and convex $k$-means show good ARI values when the dimension is equal to 12; however, they rely on continuous variables to cluster data points as their performance always increases when the proportion of continuous variables in the dataset increases. LCM and MBN rely mainly on the categorical variables to cluster data points, so they achieve good performance when the proportion of continuous variables is equal to or greater than 50% and the dimension is equal to 12. Finally, PDQ show poor clustering results in this simulation model.

In case of significant imbalance in clusters' distribution, with one cluster containing 80% of instances, the overall performance decreases as expected. On one side, when the dimension is equal to six, the general clustering performance is poor, with ARI/AMI values smaller than 0.4/0.3. On the other side, $k$-prototypes is the only method showing moderate-to-good clustering performance. Also, LCM and convex KM showed good clustering recovery when the size is large ($N = 1200$) and the ratio of continuous variables in the dataset is, respectively, low and high.

Table 3 presents the results for $K = 3$. When clusters are balanced, the overall performance improves slightly and the methods exhibit behavior consistent with the $K = 2$ scenario. In this setup, the cluster size imbalance is moderate, resulting in minimal impact on the performance. KAMILA maintains consistent clustering performance across all scenarios. As observed previously, LCM achieves high ARI values when the dataset contains a sufficient number of categorical variables. Consistent with the earlier findings, MBN is highly sensitive to variations in the sample size and the proportion of continuous variables, with these experimental factors significantly affecting the network estimation.

M3 simulation model presents a challenging scenario for all the considered methods. On one hand, neither distance-based methods nor KAMILA or LCM take into account the dependence between variables. On the other hand, the MBN assumes that the data come from a mixture of $K$ BNs rather than a single one. The results (Figure 5) are generally poor. However, in four out of eight scenarios, MBN outperforms the other methods. When $K$ increases, all methods except MBN show an improvement in clustering accuracy. As previously observed, only MBN benefits significantly from the increase in the sample size. Both KAMILA and LCM demonstrate comparable performance, with peak results achieved when $K = 4$ and $K = 6$, respectively.

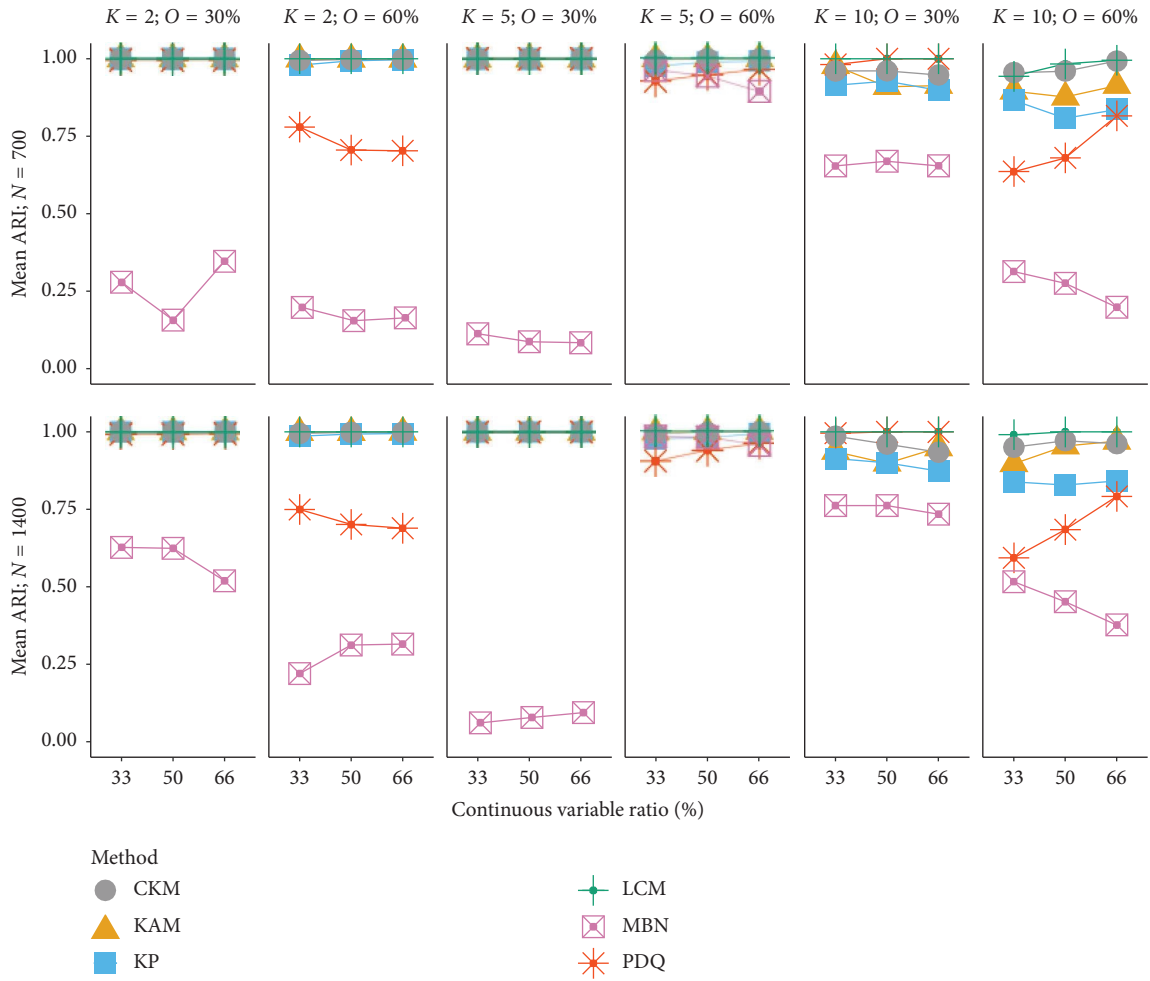

FIGURE 4: Mean ARI values from the multivariate Gaussian (M1) simulation model varying number of clusters ($K$), sample size ($N$), cluster overlap ($O$), and proportion of continuous variables in the dataset.

TABLE 2: Mean ARI/AMI values for the exponential-discrete (M2) simulation model varying size ($N$), cluster distribution ($\pi^*$), dimension ($p$), and proportion of continuous variables in the dataset.

| Cont. prop | $p = 6$ | | | $p = 12$ | | |
|---|---|---|---|---|---|---|
| | 33% | 50% | 66% | 33% | 50% | 66% |
| $\pi^* = (1/2, 1/2)$; $N = 300$ | | | | | | |
| KAMILA | 0.51/0.43 | 0.52/0.43 | 0.59/0.49 | 0.82/0.73 | 0.85/0.77 | 0.81/0.72 |
| $K$-prototypes | 0.52/0.42 | 0.47/0.38 | 0.52/0.42 | 0.72/0.61 | 0.76/0.66 | 0.74/0.64 |
| PDQ | 0.00/0.00 | 0.02/0.02 | 0.03/0.03 | 0.00/0.00 | 0.00/0.00 | 0.00/0.00 |
| Convex KM | 0.37/0.30 | 0.39/0.32 | 0.49/0.40 | 0.68/0.59 | 0.75/0.65 | 0.72/0.62 |
| MBN | 0.32/0.26 | 0.30/0.23 | 0.06/0.05 | 0.62/0.52 | 0.60/0.50 | 0.14/0.11 |
| LCM | 0.44/0.38 | 0.39/0.35 | 0.30/0.31 | 0.76/0.66 | 0.70/0.63 | 0.59/0.52 |
| $\pi^* = (1/2, 1/2)$; $N = 1200$ | | | | | | |
| KAMILA | 0.64/0.53 | 0.68/0.57 | 0.61/0.51 | 0.88/0.81 | 0.87/0.79 | 0.86/0.77 |
| $K$-prototypes | 0.48/0.38 | 0.50/0.40 | 0.53/0.43 | 0.72/0.61 | 0.76/0.66 | 0.78/0.68 |
| PDQ | 0.01/0.01 | 0.00/0.00 | 0.00/0.00 | 0.00/0.00 | 0.00/0.00 | 0.00/0.00 |
| Convex KM | 0.42/0.33 | 0.45/0.36 | 0.50/0.40 | 0.69/0.59 | 0.73/0.64 | 0.76/0.67 |
| MBN | 0.21/0.18 | 0.24/0.19 | 0.06/0.05 | 0.78/0.68 | 0.64/0.56 | 0.18/0.15 |
| LCM | 0.46/0.41 | 0.39/0.37 | 0.31/0.30 | 0.77/0.69 | 0.70/0.62 | 0.58/0.52 |
| $\pi^* = (1/5, 4/5)$; $N = 300$ | | | | | | |
| KAMILA | −0.06/0.05 | −0.02/0.09 | 0.03/0.13 | 0.11/0.19 | 0.37/0.36 | 0.37/0.34 |
| $K$-prototypes | 0.30/0.22 | 0.27/0.22 | 0.18/0.18 | 0.62/0.47 | 0.63/0.50 | 0.62/0.48 |
| PDQ | 0.00/0.00 | 0.03/0.01 | 0.00/0.00 | −0.04/0.03 | −0.02/0.01 | 0.02/0.03 |
| Convex KM | −0.04/0.06 | −0.01/0.08 | 0.02/0.11 | 0.21/0.22 | 0.42/0.35 | 0.56/0.44 |
| MBN | 0.06/0.07 | 0.03/0.05 | 0.01/0.03 | 0.22/0.20 | 0.18/0.17 | 0.04/0.06 |
| LCM | 0.13/0.15 | 0.20/0.19 | 0.00/0.07 | 0.74/0.61 | 0.72/0.57 | 0.36/0.34 |

TABLE 2: Continued.

| Cont. prop | p = 6 | | | p = 12 | | |
|---|---|---|---|---|---|---|
| | 33% | 50% | 66% | 33% | 50% | 66% |
| $\pi^* = (1/5, 4/5)$; $N = 1200$ | | | | | | |
| KAMILA | −0.01/0.10 | 0.07/0.16 | 0.13/0.18 | 0.52/0.43 | 0.56/0.45 | 0.56/0.45 |
| K-prototypes | 0.36/0.26 | 0.32/0.24 | 0.35/0.26 | 0.62/0.47 | 0.66/0.50 | 0.68/0.52 |
| PDQ | −0.01/0.00 | 0.01/0.00 | 0.01/0.02 | −0.02/0.01 | −0.02/0.01 | 0.04/0.03 |
| Convex KM | −0.04/0.06 | 0.01/0.10 | 0.08/0.14 | 0.36/0.28 | 0.52/0.39 | 0.64/0.48 |
| MBN | 0.04/0.05 | 0.02/0.03 | 0.02/0.02 | 0.26/0.22 | 0.13/0.13 | 0.05/0.08 |
| LCM | 0.04/0.12 | 0.02/0.11 | −0.04/0.04 | 0.82/0.66 | 0.74/0.59 | 0.44/0.38 |

*Note:* In this configuration, the number of clusters is $K = 2$.

TABLE 3: Mean ARI/AMI values for the exponential-discrete (M2) simulation model varying size ($N$), cluster distribution ($\pi^*$), dimension ($p$), and proportion of continuous variables in the dataset.

| Cont. prop | p = 6 | | | p = 12 | | |
|---|---|---|---|---|---|---|
| | 33% | 50% | 66% | 33% | 50% | 66% |
| $\pi^* = (1/3, 1/3, 1/3)$; $N = 300$ | | | | | | |
| KAMILA | 0.41/0.41 | 0.61/0.60 | 0.68/0.65 | 0.94/0.91 | 0.91/0.88 | 0.90/0.86 |
| K-prototypes | 0.46/0.44 | 0.55/0.54 | 0.59/0.57 | 0.72/0.68 | 0.77/0.73 | 0.78/0.74 |
| PDQ | 0.06/0.08 | 0.04/0.08 | 0.07/0.09 | 0.07/0.08 | 0.03/0.04 | 0.02/0.03 |
| Convex KM | 0.36/0.37 | 0.44/0.45 | 0.57/0.56 | 0.68/0.66 | 0.72/0.69 | 0.78/0.74 |
| MBN | 0.44/0.42 | 0.48/0.46 | 0.18/0.18 | 0.58/0.55 | 0.44/0.41 | 0.26/0.26 |
| LCM | 0.65/0.63 | 0.65/0.65 | 0.55/0.56 | 0.93/0.90 | 0.88/0.85 | 0.85/0.81 |
| $\pi^* = (1/3, 1/3, 1/3)$; $N = 1200$ | | | | | | |
| KAMILA | 0.39/0.39 | 0.74/0.69 | 0.74/0.69 | 0.95/0.92 | 0.94/0.91 | 0.92/0.89 |
| K-prototypes | 0.51/0.48 | 0.60/0.57 | 0.60/0.59 | 0.70/0.67 | 0.74/0.72 | 0.78/0.75 |
| PDQ | 0.06/0.07 | 0.09/0.12 | 0.07/0.08 | 0.06/0.07 | 0.00/0.01 | 0.01/0.02 |
| Convex KM | 0.35/0.36 | 0.50/0.49 | 0.57/0.55 | 0.70/0.67 | 0.74/0.71 | 0.78/0.75 |
| MBN | 0.29/0.29 | 0.42/0.41 | 0.08/0.08 | 0.43/0.43 | 0.40/0.40 | 0.19/0.20 |
| LCM | 0.64/0.63 | 0.66/0.64 | 0.54/0.55 | 0.94/0.90 | 0.89/0.86 | 0.83/0.80 |
| $\pi^* = (1/2, 1/4, 1/4)$; $N = 300$ | | | | | | |
| KAMILA | 0.60/0.51 | 0.72/0.66 | 0.72/0.66 | 0.91/0.86 | 0.91/0.87 | 0.92/0.87 |
| K-prototypes | 0.49/0.45 | 0.62/0.57 | 0.60/0.55 | 0.73/0.65 | 0.77/0.71 | 0.78/0.73 |
| PDQ | 0.06/0.07 | 0.13/0.17 | 0.08/0.17 | 0.09/0.11 | 0.08/0.12 | 0.01/0.13 |
| Convex KM | 0.46/0.41 | 0.55/0.53 | 0.56/0.56 | 0.66/0.63 | 0.70/0.67 | 0.77/0.73 |
| MBN | 0.42/0.40 | 0.41/0.40 | 0.24/0.22 | 0.53/0.51 | 0.41/0.39 | 0.24/0.24 |
| LCM | 0.62/0.59 | 0.58/0.58 | 0.41/0.48 | 0.90/0.85 | 0.87/0.82 | 0.83/0.80 |
| $\pi^* = (1/2, 1/4, 1/4)$; $N = 1200$ | | | | | | |
| KAMILA | 0.69/0.59 | 0.78/0.69 | 0.78/0.70 | 0.96/0.92 | 0.95/0.90 | 0.93/0.89 |
| K-prototypes | 0.55/0.50 | 0.61/0.55 | 0.61/0.58 | 0.73/0.66 | 0.76/0.71 | 0.80/0.74 |
| PDQ | 0.07/0.08 | 0.05/0.13 | 0.11/0.19 | −0.01/0.06 | 0.07/0.11 | 0.02/0.13 |
| Convex KM | 0.48/0.44 | 0.54/0.52 | 0.60/0.58 | 0.68/0.64 | 0.74/0.70 | 0.79/0.74 |
| MBN | 0.32/0.34 | 0.30/0.30 | 0.12/0.12 | 0.31/0.35 | 0.30/0.34 | 0.24/0.27 |
| LCM | 0.64/0.61 | 0.65/0.62 | 0.43/0.49 | 0.94/0.89 | 0.91/0.86 | 0.88/0.82 |

*Note:* In this configuration, the number of clusters is $K = 3$.

Finally, the last simulation model M4 explores the behavior of the considered methods when data are generated from a mixture of $K$ BNs. Compared to simulation model M3, where a precise estimation of the single network is necessary to obtain good clustering results, the data generated by this model rely more on the estimation of the parameters of each of the $K$ BN, since they have all the same network structure.

The results (Figure 5) show significant improvement for all methods compared to the results of M3, particularly for KAMILA, which outperforms the other methods in five out of six scenarios and maintains a consistently good performance throughout. As anticipated, an increase in the number of clusters has a negative impact on overall performance, with the exception of LCM, which demonstrates a notable improvement when moving from $K = 2$ to $K = 4$. As previously noted, the impact of an increase in the sample size on clustering performance is most notable in the case of MBN. The results of the last were not as good as expected. Even when the data are generated to meet the assumption of a mixture of BN, it consistently fails to estimate the network.

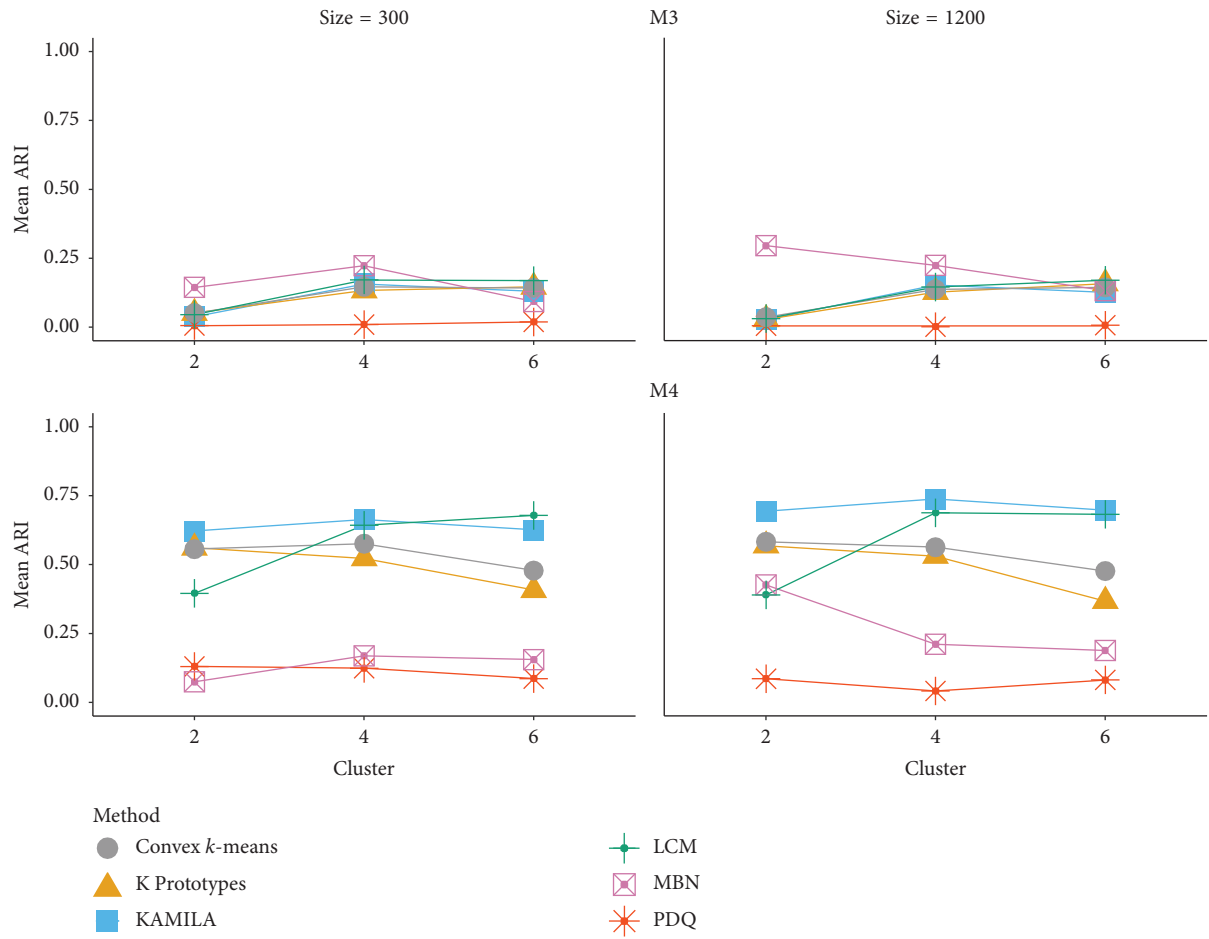

FIGURE 5: Mean ARI values for the BN classifier (M3) and mixture of BN (M4) simulation model varying the number of clusters ($K$) and sample size ($N$). The dataset contains three continuous and three categorical variables.

## 5. Discussion and Perspectives

We conducted a comparative analysis of the principal state-of-the-art algorithms for clustering mixed-type data. We performed four comprehensive simulation studies on mixed-type data with the intention of discerning the strengths and limitations of each algorithm. The aim was to gain insights into the performance of these algorithms and to identify the most suitable ones for various scenarios. By employing simulation models, we could thoroughly examine the behavior of different methods under various conditions. These conditions include scenarios where the data follow a Gaussian distribution, situations with non-normal data containing extreme values, and cases where there are significant interactions between variables and with the cluster membership.

From a clustering performance perspective, three groups of methods can be identified. The top performing group includes LCM and KAMILA. As expected and in line with Jimeno et al. [13] and Costa et al. [15], both methods performed best when clusters are spherical (M1), as they inherently favor spherical distributions. In these cases, LCM emerged as the optimal method, delivering superior results even in scenarios with significant cluster overlap. However, with large sample sizes and when the proportion of categorical variables in the dataset is at least half, LCM also demonstrated good performance on non-Gaussian distributions. KAMILA was the best-performing method in two out of four simulation models (M2 and M4) and ranked among the best in the remaining models. In cases where no prior information about data distribution or characteristics is available, a common scenario when clustering real-world data, KAMILA is the preferred choice. As noted by Foss et al. [10] and Costa et al. [15], KAMILA struggles with small sample sizes due to its reliance on estimating a multinomial model for categorical variables, a limitation shared with LCM. In addition, KAMILA was the method least affected by the ratio of continuous to categorical variables in the dataset.

The middle-performing group includes $k$-prototypes and convex KM, both exhibiting similar behavior. These methods are recommended for datasets with a higher proportion of continuous variables and small sample sizes. However, $k$-prototypes demonstrate slightly better performance than convex KM and significantly outperform all other methods when the cluster size imbalance is pronounced.

Finally, the bottom-performing group comprises PDQ and the MBN, both showing generally poor performance. PDQ performs well only when the data follow a Gaussian distribution but is highly sensitive to cluster overlap. BN-based methods, while potentially valuable when the underlying data structure is known or suspected, depend heavily on accurately estimating this structure for effective clustering. Consequently, the choice of the algorithm for network estimation becomes a critical factor influencing the performance.

We have identified two particularly challenging scenarios: when the distribution of cluster sizes is highly uneven and when there are strong interactions between variables with explicit dependence on cluster membership. In these realistic scenarios, only $k$-prototypes and MBN achieve above average ARI values; however, clustering recovery is still poor.

The current study has some limitations. The datasets generated using M1 follow a finite mixture of multivariate Gaussian distributions. Ideally, as suggested in related works, datasets should be generated from pure mixed-type data models. However, controlling cluster overlap in this scenario becomes increasingly challenging when dealing with more than two clusters. The datasets generated using M2 are based on a mixture of exponential-discrete distributions, but cluster overlap is not explicitly manipulated. While the use of the exponential distribution is motivated by real-world data from the aerospace industry, exploring other non-normal distributions would provide a broader perspective. Furthermore, ordinal variables were not considered in the study, even though they are frequently encountered in real-world datasets and often treated as continuous variables. This limitation could impact clustering performance and reduce the study's applicability to practical scenarios. Finally, in the absence of a general consensus, each method was fitted 10 times for every unique combination of experimental factors. Whenever it exists, the number of random initializations for each method was maintained at its default value.

The results highlight that the clustering analysis is highly context-dependent, varying significantly based on the specific characteristics of the problem. While this study focuses on the leading methods within the relatively limited literature on clustering mixed-type data, it does not suggest that alternative approaches should be overlooked.

All methods presented in the study belong to nonhierarchical clustering techniques (1). Among all, only MBN provides partially interpretable clustering results.

Currently, the options for selecting a clustering method suitable for mixed-type data problems are limited. In addition, the literature barely addresses relevant concerns such as the optimal number of clusters or the interpretability of the results. Integrating nonhierarchical methods such as CUBT by Fraiman et al. [35], Ghattas et al. [36], or DIV-CLUS-T [37] could effectively address both issues and expand the range of available approaches.

## Data Availability Statement

The data that support the findings of this study are openly available in GitHub at https://github.com/Alsanchez13/Clustering-approaches-for-mixed-type-data-A-comparative-study.

## Conflicts of Interest

The authors declare no conflicts of interest.

## Funding

## Endnotes

$^1$FAMD = Factor Analysis for Mixed Data.
$^2$RKM = Reduced $k$-means.